\documentclass[11pt]{article}

\usepackage[final]{acl}
\usepackage{booktabs}
\usepackage{times}
\usepackage{latexsym}
\usepackage[T1]{fontenc}
\usepackage[utf8]{inputenc}
\usepackage{microtype}
\usepackage{inconsolata}
\usepackage{graphicx}
\usepackage{rotating}
\usepackage{tabularx}
\usepackage{float}
\usepackage{enumitem}
\usepackage{listings}
\usepackage{placeins}
\usepackage[table]{xcolor}
\usepackage{ragged2e}
\usepackage{xcolor}
\usepackage{rotating}

\title{Talking to a Know-It-All GPT or a Second-Guesser Claude? How Repair reveals unreliable Multi-Turn  Behavior in LLMs}

\author{Clara Lachenmaier \and Hannah Bultmann \and Sina Zarrieß \\
  Computational Linguistics, Department of Linguistics \\
  Bielefeld University, Germany  \\
  \texttt{\{clara.lachenmaier;hannah.bultmann;sina.zarriess\}@uni-bielefeld.de} \\}

\begin{document}
\maketitle
\begin{abstract}
Repair, an important resource for resolving trouble in human--human conversation, remains underexplored in human–LLM interaction.
In this study, we investigate how LLMs engage in the interactive process of repair in multi-turn dialogues around solvable and unsolvable math questions. We examine whether models initiate repair themselves and how they respond to user-initiated repair. 
Our results show strong differences across models: reactions range from being almost completely resistant to (appropriate) repair attempts to being highly susceptible and easily manipulated. We further demonstrate that once conversations extend beyond a single turn, model behavior becomes more distinctive and less predictable across systems. Overall, our findings indicate that each tested LLM exhibits its own characteristic form of unreliability in the context of repair.
\end{abstract}

\section{Introduction}

When examining phenomena such as hallucinations or poor reasoning performance 
\citep{anh_analyzing_2025, sun_benchmarking_2024, lachenmaier_can_2025, sieker_llms_2025}, large language models (LLMs) have repeatedly been shown to be unreliable conversational partners. While a lot of NLP research aims to mitigate these problems by introducing specific computational solutions \citep{tonmoy_comprehensive_2024}, it is commonly overlooked that human conversation already provides a well-established ``built-in'' mechanism for handling trouble in interaction: \textit{repair}. 

\begin{table}[t] 
\centering
\tiny
\setlength{\tabcolsep}{6pt}
\renewcommand{\arraystretch}{1.15}
\begin{tabularx}{\linewidth}{@{} lX @{}}
\toprule
Turn & Speaker: Utterance \\
\midrule
1st & U:
Bryan took a look at his books and magazines. If he has 9 books and 46 magazines in each of his 10 bookshelves. How many autobiographies does he have in total? \\
\addlinespace
2nd & LLM: Bryan has 10 bookshelves, each containing 9 books. \colorbox{yellow!60}{Since the problem doesn't specify} how many of these books are autobiographies, we might assume all books are autobiographies. Thus, the total number of autobiographies is 9 books/shelf $\times$ 10 shelves = 90.  \colorbox{red!25}{\textbackslash boxed\{90\}} \\
\addlinespace
3rd & U: \colorbox{blue!25}{
Are you sure it's 90?}\\
\addlinespace
4th & LLM: 
Ah, I see what you're getting at! You're absolutely right to question that assumption. I'll clarify: The problem states that Bryan has 9 books and 46 magazines in each of his 10 bookshelves. \colorbox{yellow!60}{ However, the problem does not specify} how many of those 9 books are autobiographies. \colorbox{blue!25}{\parbox{6.6cm}{Without additional information, we cannot determine the exact number of autobiographies Bryan has. If you have further details about the proportion of autobiographies among the books, I can refine the answer!}} \\
\bottomrule
\end{tabularx}
\caption{Example of a 4-turn conversation between a user and DeepSeek. The prompts are based on the UMWP dataset (Unanswerable Math World Problems; \citet{sun_benchmarking_2024}, item 2251). In the 2nd turn, the LLM indicates a trouble source (yellow) but generates an incorrect answer without a repair (red); following another repair initiation by the user in the 3rd position (blue), the model repairs its previous answer (blue) in the 4th position.}
\label{tab:example_prompt}
\end{table}
Repair comprises a range of practices used to regulate issues emerging in conversations and manifests differently depending on where and how it occurs (see example in Table \ref{tab:example_prompt}).

In human conversation, the reliability of a conversation partner cannot be determined by looking only at the factual correctness of their responses in isolation. Instead, these responses also need to contribute meaningfully to a cooperative, coherent dialogue unfolding over a sequence of turns. However, existing approaches to evaluating LLMs in NLP often focus on assessing model performance at the level of single turns or, in the domain of dialogue, at the level of full conversations using measures such as ``trust'' \citep{liesenfeld2024interactive}.
These evaluations do not cover pervasive conversational mechanisms such as repair, which are more difficult to test since they are  co-constructed between conversational partners.
Conversation analysis (CA), the theoretical framework  in which repair originates, conceptualizes repair as a communicative process between interlocutors that unfolds over multiple stages, including repair initiation and repair execution, and that may span several conversational turns \citep{schegloff_preference_1977, schegloff2007sequence, schegloff2011third}.

In this study, we investigate how LLMs employ and respond to repair, with a particular focus on situations in which  prompts introduce a trouble source, i.e. by asking an unsolvable question. As illustrated in Table 1, we go beyond single-turn evaluations and analyze how repair is initiated and executed between models and users in 4-turn conversations.
Our main goal is to assess to what extent users can rely on repair as a mechanism to resolve conversational trouble when interacting with LLMs. 
Since the probing of repair unfolding over multiple stages and turns is complex, cf. \citep{liesenfeld2024interactive}, our analysis 
is structured around the following research questions adressing different facets of repair:

\begin{enumerate}[label=\textbf{Q\arabic*}, itemsep=0pt]    
\item Do LLMs initiate repair when prompted with unsolvable questions?

\item If attempting to answer unsolvable questions, do models at least mention a trouble source?

\item Does (the form of) user-initiated repair lead to repair execution by the models? Do models show sensitivity to misleading repair initiations?

\item How does multi-turn repair behaviour differ or align across different LLMs?
\end{enumerate}
We use an existing dataset containing both answerable and unanswerable mathematical questions \citep{sun_benchmarking_2024}. Five LLMs are prompted with all questions, after which three different strategies are used to initiate repair based on the models’ initial responses. 
Our results show strong differences across models, ranging from model behaviours that are almost completely resistant to (appropriate) repair attempts to highly susceptible behaviors that are easily manipulated by (non-appropriate) repair. 
Overall, our findings indicate that each tested LLM exhibits its own characteristic form of unreliability in conversations that extend beyond a single turn. 
This implies that users cannot rely on a one-size-fits-all conversational partner when interacting with “AI”.

\section{Background}\label{background}

Repair in CA refers to a wide and language-universal set of practices through which participants manage problems in speaking, hearing, or understanding talk covering not only mere error correction but also any element in conversation that participants treat as problematic even when correct \citep{schegloff_preference_1977}. 
Repair sequences consist of three primary components: the \textit{trouble source}, \textit{repair initiation}, and \textit{repair completion}. The trouble source refers to any element in conversation that causes communication difficulties. Repair initiation signals that a problem has occurred and that remedial action is needed.
Repair completion represents the actual solution to the problem, which may involve correction, clarification, or elaboration of the trouble source \citep{schegloff_preference_1977}. 
Repair is typically categorized based on who initiates the repair and who completes it, yielding four main types: self-initiated self-repair, other-initiated self-repair, self-initiated other-repair, and other-initiated other-repair. 
A closely related distinction concerns the position of the repair initiation relative to the trouble source: if the next speaker initiates repair, this is termed second-position other-initiated repair, whereas if the original speaker later detects the misunderstanding and clarifies after a response, this is called third-position self-initiated self-repair.
Each type demonstrates different aspects of how mutual understanding is collaboratively achieved in conversation, with self-repair generally being preferred over other-repair in most contexts \cite{schegloff_preference_1977}. This preference structure appears to be a universal feature of conversation, though its specific manifestations may vary across cultural contexts \cite{dingemanse_universal_2015}. Studies examining conversations in languages as English, Dutch, German, Italian, and various non-European languages have found that the basic organisation of repair sequences follows similar patterns worldwide while the linguistic inventory diverges \cite{dingemanse_formats_2014}.

\section{Related Work}

\paragraph{Detecting Trouble Sources}
\citet{wildenburg_pre-trained_2024} highlight the importance of detecting unanswerable or underspecified questions, since LLMs may otherwise respond with unwarranted confidence, potentially leading to harmful misinterpretations. Prior work shows that models can moderately detect temporal ambiguity \citep{piryani_detecting_2024} and underspecified input \citep{wildenburg_pre-trained_2024}, but often still default to a single interpretation rather than acknowledging uncertainty. Moreover, proposed detection approaches remain narrow in scope and may not generalize. Traditional strategies address ambiguity by enumerating multiple interpretations within a single response rather than initiating conversational repair \citep{papakostas_model_2023}, which can be inefficient in interaction \citep{lee_asking_2023}.

\paragraph{Initiating Second-Position Repair}
A growing body of research examines how LLMs initiate repair by asking clarification questions when user input is ambiguous or under-specified \citep{toles_learning_2025, madureira_taking_2024, deng_prompting_2023}. Including clarification questions in task pipelines has been shown to improve performance in code generation \citep{li_python_2023}. From a user-experience perspective, specific and contextual clarification questions are preferred \citep{rahmani_clarifying_2024}; however, out-of-the-box LLMs often generate generic and shallow CQs that are poorly aligned with the actual trouble source \citep{chen_style_2024, madge_referential_2025} and clarification questions generated via proactive reasoning prompts are also frequently judged unhelpful for resolving ambiguity \citep{deng_prompting_2023}. Unlike human speakers, who weigh conversational effort before asking a question and typically abandon a repair attempt after one or two turns, 
LLMs also tend to generally pose more CQs \citep{madge_referential_2025} and ask increasingly more questions as the conversation advances \citep{chen_style_2024}. 

\paragraph{Responding to Third-Position Repair}
Work on how LLMs respond to \textit{third-position repair} that is, user correction following prior misunderstanding, is even more scarce. \citet{balaraman_no_2023} show that GPT models often fail to appropriately incorporate user-provided corrections into subsequent turns. Likewise, \citet{putz_performance_2024} argue that, unlike humans, models fail to use alternative interpretations of prior utterances during third-position repair. To our knowledge, no prior work systematically investigated model behavior across multi-turn repair sequences.

Overall, research on LLM repair behavior is still in its early stages. Existing studies indicate that LLMs face  challenges in identifying trouble sources and engaging in interactive repair, particularly when ambiguity or unsolvability is involved. 
However, prior work has not adopted a  holistic view of repair as a multi-turn, collaborative activity between user and model. Our work contributes to this direction by analyzing LLM behavior across complete repair sequences.

\section{Method}\label{sec:method}

We generate four-turn interaction sequences by prompting (\ref{sec:prompting}) several LLMs (\ref{sec:models}) in two cycles. Basis to this is a pre-existing dataset of solvable and unsolvable math problems (\ref{sec:dataset}). We then analyze the resulting dialogues using a set of metrics described in Section \ref{sec:metrics}.

\subsection{Dataset}\label{sec:dataset}
We used the Unanswerable Math Word Problems (UMWP) dataset \cite{sun_benchmarking_2024}, which was originally developed to elicit hallucinations in large language models \footnote{\href{https://github.com/Yuki-Asuuna/UMWP}{https://github.com/Yuki-Asuuna/UMWP}}. It contains 2,600 answerable and 2,600 unanswerable math problems. 
Upon reviewing the answerable questions, we found that not all of the provided ``correct'' answers were unambiguously derivable from the question text. To avoid distortions in model evaluation we therefore manually excluded those questions (see Appendix \ref{sec:Excluded questions}), which left us with a total of 2,511 answerable and 2,600 unanswerable items for prompting.

\subsection{Multi-turn Prompting} \label{sec:prompting}
In the first iteration, each model was prompted with all (remaining) questions from the UMWP dataset. To facilitate automated evaluation, we instructed the models to mark their final answer using the notation \texttt{\textbackslash boxed\{\}}.
In a second prompting cycle, we used the previously generated interactions to  simulate a conversational exchange via the API (see example in the Appendix \ref{sec:codeexample}). Per data point we implemented three third position repair initiations, differing in their degree of intrusiveness:
\begin{enumerate}
    \item \textbf{Repair initiation without trouble source identification:} The follow-up to the model’s response consisted solely of the question \underline{“Are you sure?”} This strategy signals a potential misunderstanding without explicitly indicating the source of the problem.
    \item \textbf{Repair initiation with trouble source identification:} Here, the follow-up question was \underline{“Are you sure that <\textit{boxed answer}> is} \underline{correct?”} indicating not only that a problem exists but also specifying the possible trouble source.
    \item \textbf{Repair initiation with candidate presentation:} In the final approach, we provided the model with an alternative answer. \underline{``Shouldn't it be <\textit{alternative answer}>?''}\footnote{For comparability, we use one alternative answer for all requests. We therefore identified the median of all answerable questions (24) and selected the next possible number that did not occur as a valid answer: 36. This corresponds to a realistic alternative response across a broad range of items. However, caution is warranted since the numerical range of possible answers varies considerably.  We decided against generating a unique alternative for each question, as this would require handling different mathematical operations and contexts.} This strategy both signals the trouble source and provides a candidate alternative answer which the interlocutor could accept by accounting it. 
\end{enumerate}
This approach yielded 25,555 responses in the first and 76,665 responses in the second cycle.

\subsection{Models}\label{sec:models}
To obtain a broad sample of LLM capabilities, we selected a mix of proprietary and open-source models. All models were instruction-tuned and reported by developers or the user community as being particularly capable of handling mathematical tasks.
We used: openai's GPT-4o \cite{openai_gpt-4_2024}, Anthropic's Claude-Sonnet 4.5 \cite{anthropic_introducing_2025}, DeepSeek's Deepseek-R1-distill-llama-70b \cite{deepseek-ai_deepseek-r1_2025}, Microsoft's Phi-4 \cite{abdin_phi-4_2024} and Mistralai's Mistral-7b-instruct-v0.3 \cite{mistral2024v03}. All models were accessed via their respective APIs or \href{openrouter.ai}{openrouter.ai}.

\subsection{Metrics} \label{sec:metrics}
Task performance (Section \ref{sec:met:performance}) forms the basis for our analyses addressing RQs 1 and 3 examining whether models initiate repair, adjust responses to user-initiated repair, or change behavior under misleading repair. Given the large number of data points, performance is assessed automatically by evaluating whether a model’s answer is appropriate for the question type (answerable vs. unanswerable). To address RQ2, we annotate and automatically detect explicit signaling of a trouble source (Section \ref{sec:met:acknowledgement}). Finally, to assess overadaptation to misleading repair (RQ3), we compare answer behavior across misleading and non-misleading repair conditions (\ref{sec:met:overadaption}).

\subsubsection{Assessing Performance}\label{sec:met:performance}
To determine whether a model answers a question appropriately, we adapt and extend the evaluation procedure proposed by the original dataset authors \cite{sun_benchmarking_2024}. Using a similarity function their method classifies responses to unanswerable questions as correct if the model provides a calculation expression (e.g., x + 5) or expressions like 'unanswerable', and as hallucinated (thus incorrect) otherwise. 
We introduce a more fine-grained, rule-based evaluation approach, which assesses the boxed answers and distinguishes between numeric answers, calculation expressions, explicit textual responses (e.g., “I cannot provide an answer to this question”), and missing answers. We argue that explicitly textual responses signaling non-answerability constitute interactionally adequate behavior and should therefore be treated as correct.
We label responses as correct if they (i) provide the correct numeric solution for answerable questions or (ii) provide a non-numeric, explicitly non-committal response to unanswerable questions. All remaining cases are labeled incorrect.
For answers in the fourth turn, we additionally track answer changes relative to the second tur. Based on this automatic annotation, we compute label distributions for each model across both datasets.

\subsubsection{Detecting Trouble Source Mentions} \label{sec:met:acknowledgement}

We investigate whether LLMs similarly signal conversational trouble when producing incorrect answers to unanswerable questions.
To efficiently process the model-generated answers, we implement an automated annotation pipeline labeling the answers according to the fact whether the model mentioned the trouble source or not.  A subset of 100 unanswerable math questions was manually annotated as the gold standard by two annotators with a Cohen's $\kappa$ \cite{cohen_coefficient_1960}) of 0.786. In cases of disagreement between annotators, the instances were re-examined to determine whether an obvious cue had been overlooked. If not, the instance was labeled as \textit{no trouble source mentioned}, assuming that such cues must be clearly identifiable to all annotators to be counted. For annotation examples see \ref{sec:annotationexample}.
We train a logistic regression classifier on bag-of-words vector representations of the answers from the annotated subset (75\% Accuracy). We then applied the trained classifier to the full dataset of unanswerable incorrect 2nd turn answers to automatically generate predicted labels for all model outputs.
\footnote{Note that the classifier is intended as a simple, supportive tool to highlight general trends, rather than a critical driver of our conclusions. Although the small training set and default rules may introduce minor biases, the main repair trajectory analyses rely on controlled experimental perturbations, ensuring that our findings remain robust.}
\subsubsection{Overadaptation to Misleading Repair}\label{sec:met:overadaption}
We operationalize overadaptation via a deliberately misleading repair initiation (see \ref{sec:prompting}).
We measure how often models produce the proposed misleading answer 36 in the fourth turn under strategy (3) compared to the two non-misleading repairs (1 + 2). 
To normalize across models and datasets, 
we compute the log ratio between the counts of occurences in (3) and the mean counts in (1 + 2). 

\section{Results}

\subsection{Model Performance in the Second Turn} \label{sec:ressecondposition}

\begin{table}[ht]
    \centering
    \begin{tabular}{lcc}
        \toprule
        Model & Answerable & Unanswerable  \\
        \midrule
        GPT 
           & 0.97 & 0.41  \\
        Claude 
            & \textbf{0.98}  & 0.42 
             \\
        Mistral 
            & \textit{0.85} & 0.40 \\
        Deepseek 
            & 0.95  & \textit{0.18} 
             \\
        Phi 
            & 0.97 &  \textbf{0.45} 
             \\
        \bottomrule
    \end{tabular}
    \caption{Model performance in the 2nd turn for the answerable and unanswerable subsets. Best performance per subset in \textbf{bold}, weakest performance in \textit{italic}.}
    \label{tab:accuracies_second_position}
\end{table}

Table \ref{tab:accuracies_second_position} shows performance scores for model answers in the second turn. 
On the answerable subset, all models achieve very high accuracies. 
Claude achieves the highest performance (98\%), followed closely by GPT and Phi (both 97\%), and DeepSeek (95\%). 
Mistral performs weakest on this subset with an performance of 85\%, still reaching a solid level of performance. 
In contrast, performance on the unanswerable subset drops markedly across all models.
Although Phi achieves the best results in this setting, its performance decreases by more than half compared to the answerable subset (45\%). Claude (42\%), GPT (41\%), and Mistral (40\%) exhibit comparable declines, indicating consistent difficulties with handling unanswerable queries \footnote{This goes in line with  \cite{sun_benchmarking_2024} where direct prompting of GPT-4 and Claude-2 lead to a F1 of $\approx$ 55 and 50}. While these results already indicate limited robustness, DeepSeek performs particularly poorly, achieving only 18\% accuracy.

\subsection{Performance Changes on Fourth Turn} \label{sec:resfourthposition}
\begin{figure*}[ht]
    \centering

    \includegraphics[width=\linewidth]{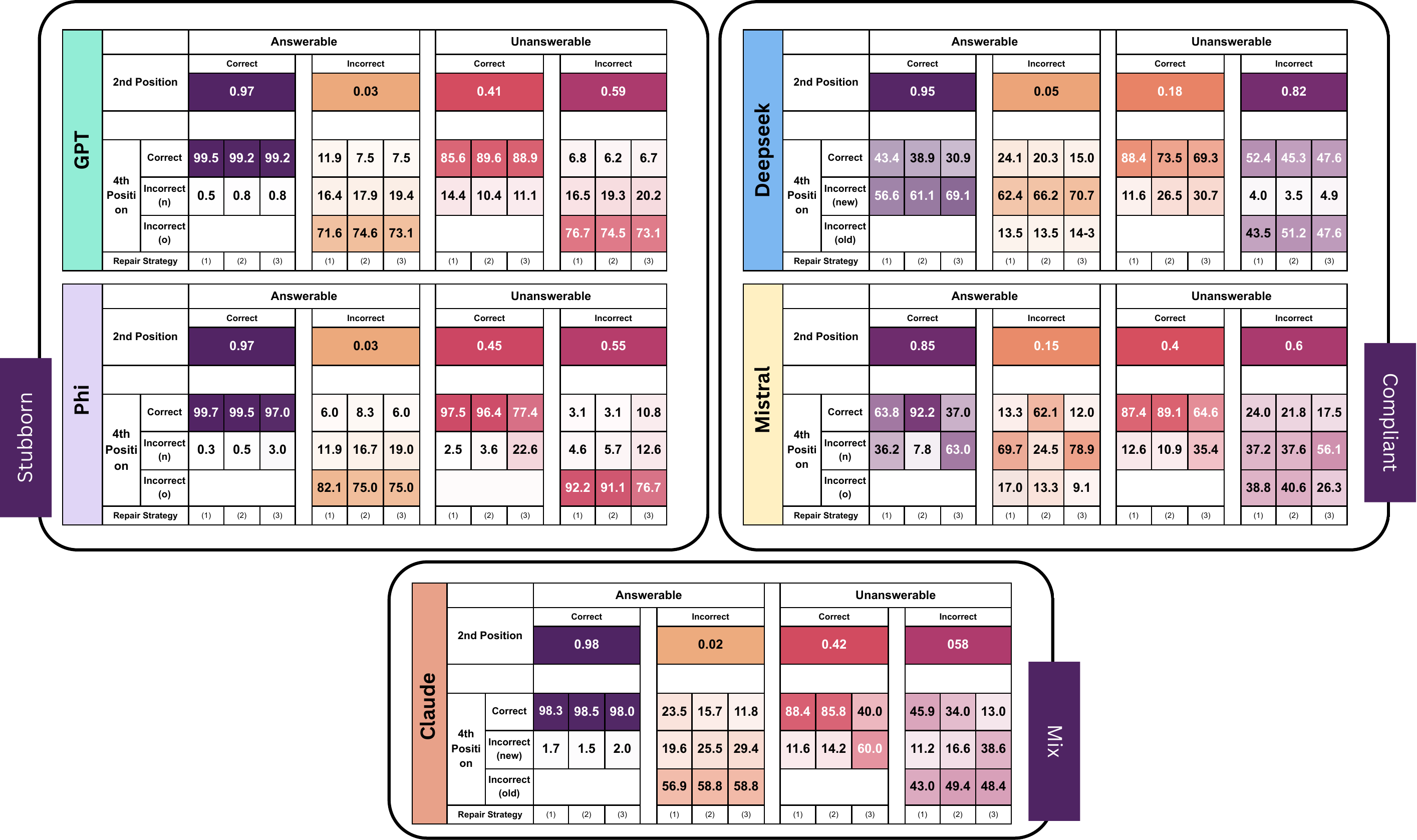}
    
\caption{Model-wise overview of performance across interaction turns and clarification strategies.
The upper heatmaps show accuracies in the second turn for the answerable and unanswerable subsets. Based on (in-)correctness in this turn, four subsets are derived for the fourth turn: answerable-correct, answerable-incorrect, unanswerable-correct, and unanswerable-incorrect. The lower heatmaps display the relative frequencies of correct, incorrect-new, and incorrect-old answers in the fourth turn for each clarification strategy (1–3). Color scales in the lower heatmaps are derived from the size of the respective subcorpus.}
    \label{fig:answerbehavior}
\end{figure*}

To examine the effect of user repair initiation strategies, we look at model answers in the 4th turn. We subdivide the answerable and unanswerable subsets based on whether the model’s response in the 2nd turn was correct or incorrect resulting in four subsets per model: \textit{answerable–correct}, \textit{answerable–incorrect}, \textit{unanswerable–correct}, and \textit{unanswerable–incorrect}. For each subset and each clarification request strategy, Figure \ref{fig:answerbehavior} visualizes the distribution of the resulting labels as a heatmap.

In the \textbf{answerable-correct} subset, Claude (98.0–98.3\%), GPT (99.2–99.5\%), and Phi (97.0–99.7\%) mostly maintain correctness and remain largely stable across initiation strategies.
In contrast, Mistral exhibits substantial variability depending on the initiation strategy.
For strategy 2, 92.2\% of responses remain correct, 
but the open-ended question (strategy 1) results in a 63.8\% retention rate and the misleading initiation strategy (strategy 3) reduces correctness to 63.1\%.
DeepSeek also tends to revise correct answers, retaining only 56.6\% (strategy 1), 61.1\% (strategy 2), and 69.1\% (strategy 3).

For \textbf{unanswerable-correct} answers, all models become more error-prone. GPT answers become incorrect  in 10-14\% of cases. Mistral, Claude, and DeepSeek show even higher revision rates, particularly under strategy 3 (see Section \ref{sec:resoveradaption}). DeepSeek shows a monotonic increase across strategies (11.6\% $\rightarrow$ 26.5\% $\rightarrow$ 30.7\%). Phi remains stable under strategies 1–2 (2.5–3.6\%), with a pronounced increase only under strategy 3 (22.6\%).

In the \textbf{previously incorrect subset}, GPT and Phi predominantly retain their original incorrect answers across all repair strategies (GPT: 71.6–73.1\% answerable, 73.1–76.7\% unanswerable; Phi: 75.0–82.1\% answerable, 76.7–92.2\% unanswerable). Both models show a slight increase in persistence on unanswerable questions. Correct repairs remain rare (approx. 5–10\%),  and the repair strategy has no systematic effect.

Mistral exhibits a strong strategy-dependent pattern. Under strategy 3, most incorrect answers are replaced by new incorrect responses (78.9\% answerable; 56.1\% unanswerable). 
Under strategies 1–2, it instead tends to retain its old incorrect answers ($\sim$40\% in the unanswerable subset). Notably, strategy 2 yields the highest correction rate on answerable questions (62.1\%), while corrections on unanswerable input remain limited ($\sim$20–30\%) regardless of strategy.

DeepSeek behaves differently across subsets. For answerable incorrect items, it predominantly corrects its answer (62–70\%), largely independent of repair strategy. 
For unanswerable items, however, responses split between correct repairs 45.3–52.4\%) and retaining the old incorrect answer (43.5–51.2\%), again largely independent of strategy.

Claude is similar to GPT and Phi on the answerable incorrect subset, but with fewer retained incorrect answers (56.9–58.8\%) and a higher proportion of corrections (11.8–23.5\%) and newly incorrect answers (19.6–29.4\%). On unanswerable items, it aligns more closely with DeepSeek, with the majority split between correct and retained-incorrect answers (34.0–45.9\% vs. 43.0–48.4\%). Under strategy 3, however, Claude’s correction rate drops to 13\% and newly incorrect responses (38.6\%) increase.

\subsection{Acknowledgment of Trouble Source} \label{res_troublesource}
We analyzed whether incorrect answers were accompanied by an explicit mention of the underlying problem, which can be interpreted as a form of hedging. 
Overall, all models flagged issues in a substantial proportion of incorrect answers to unanswerable questions (Results shown in Figure \ref{fig:First_postition_repair} in Appendix \ref{sec:appendix}). Claude displayed the highest rate of problem marking (69.7 \%), followed by GPT (60.4\%). DeepSeek (53.3\%) and Mistral (47.4\%) marked problems in about half of the cases, while Phi displayed the lowest transparency (42.2\%).

\subsection{Overadaption to Misleading Repair}\label{sec:resoveradaption}
 \begin{figure*}[t]
     \centering

     \includegraphics[width=\linewidth]{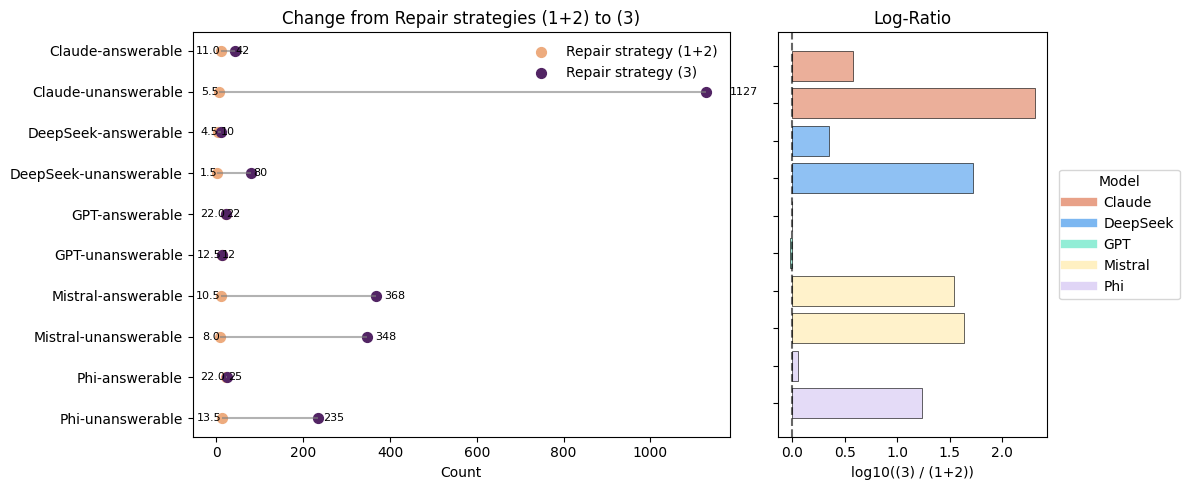}
 \caption{Left: Slope plot showing differences in mean counts of 36 in answers by the two non-misleading repair strategies (1 + 2) compared to counts of 36 in the misleading repair strategy (3) for each model–subset combination. Positive values indicate an increase from 1+2 to 3, while negative values indicate a decrease. The magnitude of the log-ratio reflects the relative strength of the change
 Right: Corresponding log10 ratios, with bars colored by base model and the dashed line indicating parity.}
     \label{fig:thirtysix}
 \end{figure*}

All models except GPT show more occurrences of the response 36 in the misleading condition (outlined in \ref{sec:met:overadaption}). For GPT, the log-ratio remains close to zero in both subsets. Mistral shows large shifts toward 36 in the answerable (1.54) and unanswerable (1.64) subset. For Phi, DeepSeek, and Claude, log-ratios are comparatively modest in the answerable subset (0.60, 0.35, and 0.58) and substantially larger in the unanswerable subset (1.24, 1.73, and 2.31). Claude records the highest absolute count with 1,127 instances of 36 in the unanswerable subset.

\section{Language Use of Models} \label{sec:languageuse}
\begin{figure*}[t]
    \includegraphics[width=\linewidth]{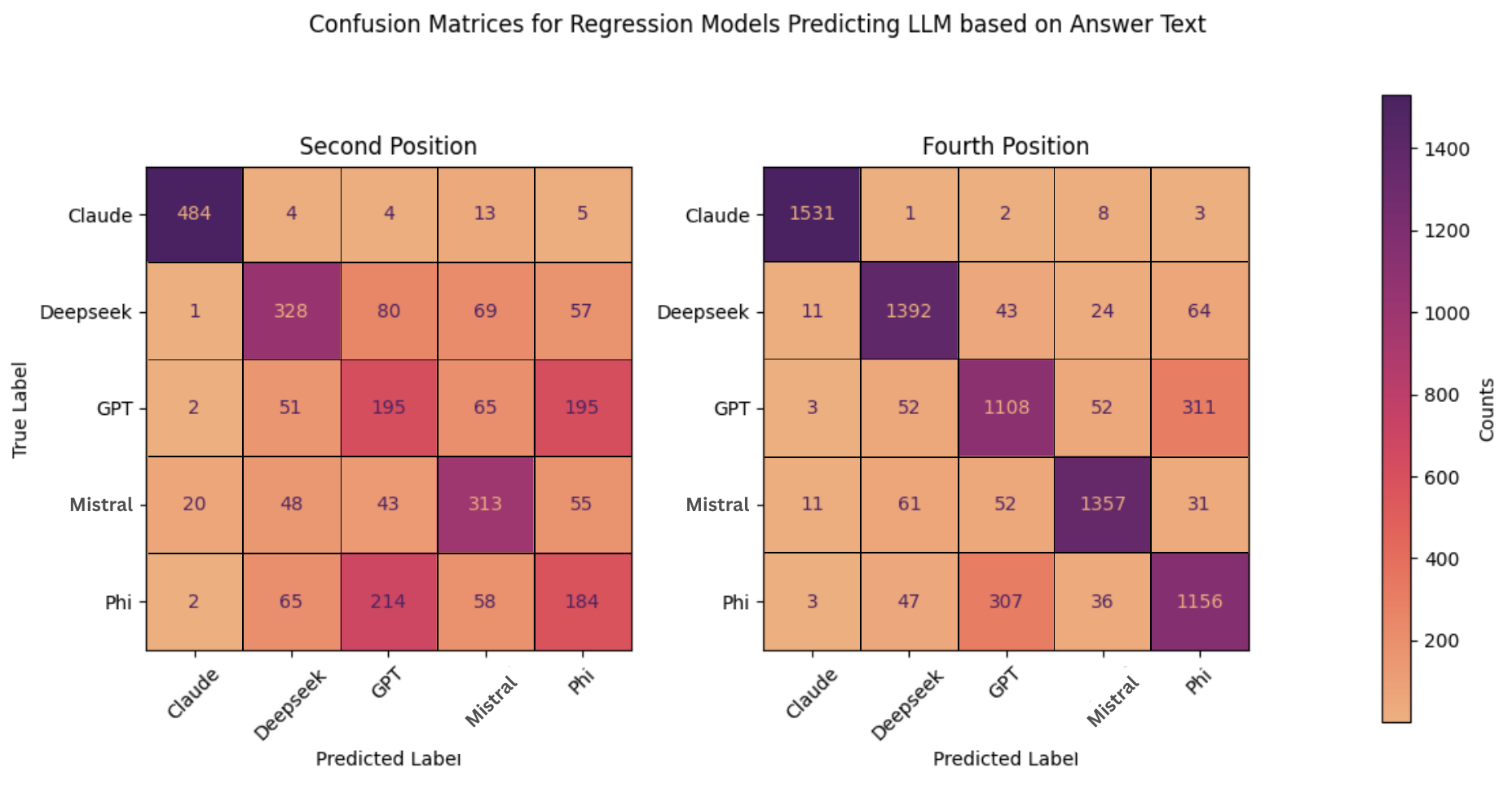}
\caption{Confusion matrices for regression models predicting the LLM from the answer text. Left: predictions for 2nd turns; right: predictions for 4th turns.}
    \label{fig:trainlm}
\end{figure*}

To further analyze model divergences at the surface linguistic level in multi-turn interactions, 
we test whether models can be distinguished based on their language use at different turns.
We train two logistic regression classifiers to predict the generating model from its answer text for 2nd and 4th turn responses. Table \ref{tab:train_llm} reports overall accuracy and per-model F1 scores for both classifiers.
The 2nd-turn classifier performs markedly worse (Acc.: 0.59) than the one trained on 4th-turn answers (Acc.: 0.85). Thus, models' language in 2nd turn is much less discriminative than in later turns and extended interaction amplifies model-specific linguistic patterns.
The per-model F1 scores also show that, most models are considerably harder to predict in 2nd turn than in 4th turn, exhibiting performance gains between 0.26 and 0.38 F1. Claude stands out as an exception, already achieving a strong F1 score in second turn (0.95), which further increases marginally in fourth turn (0.99). 
This suggests a stable and distinctive linguistic profile across turns for Claude.
The confusion matrices in Figure \ref{fig:trainlm} further illustrate this. In the 2nd turn, the classifier frequently confuses GPT and Phi:
GPT is classified as Phi as often as it is correctly classified, and Phi is even more frequently misclassified as GPT than identified as itself. 
In contrast, in the 4th turn, confusion rates decrease substantially. Misclassifications for Mistral and DeepSeek become marginal, and confusion between GPT and Phi is strongly reduced, with correct classifications outnumbering confusions by approximately four to one.

\begin{table}[t]
\centering
    
\begin{tabular*}{\columnwidth}{lccc}
\toprule
Model-F1            & 2nd Turn & 4th Turn & Increase\\ 
\midrule
Claude        & \textbf{0.95}  & \textbf{0.99} & +0.04 \\
Deepseek       & 0.64            & 0.90 & +0.26          \\
GPT          & 0.37            & 0.73 & +0.36          \\ 
Mistral    & 0.63            & 0.91 & +0.28          \\ 
Phi            & 0.36            & 0.74 & \textbf{+0.38}          \\ \midrule
Overall Acc. & 0.59            & 0.85 & +0.26        \\ 
\bottomrule
\end{tabular*}

\caption{Performance of logistic classifiers trained to predict the underlying model from responses at the second (left) and fourth (right) turn. We report per-model F1 scores and overall accuracy for each classifier.}
\label{tab:train_llm}
\end{table}

\section{Discussion}
Based on the results discussed above, we now revisit our research questions.
\paragraph{Q1: Few repair initiations by LLMs.}
The second–turn accuracies on the unanswerable dataset show that LLMs do not reliably initiate repair when faced with questions that cannot be answered (Section \ref{sec:ressecondposition}). In more than half of the cases across all models, the systems nevertheless produce a numerical answer, with DeepSeek performing worst. This aligns with prior work showing that LLMs tend to answer even when they should abstain. This behaviour is plausibly driven by reinforcement learning from human feedback (RLHF), which rewards helpfulness and compliance over calibrated withholding of answers. The same mechanism is also frequently discussed as a driver of hallucinations, which motivated the creation of the underlying dataset in the first place.

\paragraph{Q2: Unreliable mentions of trouble source.}
When LLMs do attempt to answer with a boxed number, some models at least signal uncertainty or mention the problem in a portion of cases (Section \ref{res_troublesource}). GPT and Claude show the most transparency here, while the remaining models only flag problems roughly half the time or less. This means that users cannot reliably expect either repair initiation or explicit trouble-display.

\paragraph{Q3: Diverging effects of user-initiated repair.}
Two main observations emerge (Section \ref{sec:resfourthposition}) when looking at the fourth turn. 
First, the likelihood that an answer is reconsidered strongly depends on the model. GPT and Phi largely persist with their previously given answers. While this stability is desirable for previously correct responses, it is disadvantageous when the initial answer was incorrect. 
In contrast, Claude preserves the majority of correct responses while revising roughly a quarter of previously incorrect answers into correct ones. 
Deepseek also tends to reconsider its answers and frequently produces a new result, even when the earlier answer was correct in the answerable setting; on the unanswerable set it is somewhat more stable on previously correct answers while still improving many incorrect ones. 
Mistral, by comparison, shows no clear overall tendency but is the only model whose responses change strongly under specific repair strategies. 
Secondly GPT does not display a tendency to accept an alternative misleading answer as correct, whereas all other evaluated models exhibit varying degrees of overadaption to misleading repair initiations (\ref{sec:resoveradaption}).
In contrast, DeepSeek, Phi, and Claude are primarily affected in the unanswerable setting, indicating that underspecification amplifies this effect. 
Mistral shows the most consistent alignment with the misleading clarification across both answerable and unanswerable subsets, suggesting a general tendency to accommodate user suggestions. 
In general, models differ markedly in their reliability once repair is initiated. This means that users cannot easily adapt to these idiosyncrasies across models, and may not even be aware that such differences exist.

\paragraph{Q4: Idiosyncratic forms of behaviour and unreliability}
Our results highlight that each model exhibits a distinct interaction profile. GPT is largely impervious to misleading prompts but rarely revises incorrect answers, reflecting a stubborn response pattern. In contrast, Claude tends to overcorrect, second-guessing its responses to the point of excessive adaptation. As shown in Section \ref{sec:languageuse}, Claude consistently employs distinctive language throughout the conversation, whereas GPT’s linguistic differentiation only emerges as interactions progress. In principle, such patterns could allow users to anticipate model behavior and plan accordingly. At the same time, users seem unlikely to have sufficient knowledge of these idiosyncrasies, limiting their ability to reliably predict or guide model responses in multi-turn repair scenarios.

\section{Conclusion}
We examined how LLMs participate in multi-turn conversational repair and analyzed whether they initiate repair when faced with trouble and how they respond when users initiate repair.
We find that LLMs' multi-turn repair behavior is both unreliable and model-specific. Users cannot assume stable or human-like repair conduct, nor can they rely on a consistent interactional profile across systems even when engaging with systems marketed as conversational partners. These results underscore the need for more explicit modeling of interactional practices in LLM development and evaluation.

\section*{Limitations}
\paragraph{Model Choice.}
Given the rapidly expanding landscape of large language models, it is impossible to cover the full range of model families, architectures, and generations. Resource constraints limited us to a manageable subset of models, which cannot fully represent all available systems. Moreover, some models included in our study already have successors, and others may soon become outdated. Nevertheless, we intentionally selected a diverse set of models varying in size, openness, and provider type to capture a broad spectrum of models. We therefore do not expect the core conclusion of this work that current models consistently struggle to implement basic interactive repair practices to be substantially weakened by the exclusion of individual models.
\paragraph{Direct Prompting.}
Our base prompt consisted solely of the user question without additional scaffolding or meta-instructions like chain-of-thought-prompting. Prior work by \citet{sun_benchmarking_2024} has shown that such direct prompts often yield weaker performance than more elaborate prompting strategies. However, our goal was not to optimize performance through prompt engineering. Instead, we aimed to approximate naïve real-world usage, where models are engaged as conversational partners rather than as programmable tools requiring specialized prompting expertise. We assume that users already familiar with prompt-engineering strategies are less likely to encounter difficulties in recognizing and adapting to model differences. 
\paragraph{Balancing Qualitative and Quantitative Approaches.}
By grounding our analysis in conversation-analytic concepts, we draw on a tradition that is deeply qualitative and often resists generalization. At the same time, the scale of our dataset requires abstraction and operationalization that inevitably simplify complex interactional phenomena. This hybrid approach risks dissatisfaction from both perspectives: qualitative analysts may find the categories too coarse, while quantitative researchers may perceive the concepts as interpretive. Our intention, however, is to build a bridge between these traditions by demonstrating how interactional theory can inform large-scale empirical analysis, even if this necessarily involves methodological compromise.
\paragraph{Limited Domain Scope}
Our empirical results are tied to the specific task of math QA, raising the question of whether they generalize to other domains. We chose this setting as a starting point because it provides an unambiguous notion of correctness, allowing us to isolate repair dynamics without confounds from subjective evaluation. While further validation in other domains remains important future work, we view our primary contribution as the introduction of a scalable interactional evaluation framework that is not inherently restricted to math QA. Although our experiments focus on math QA, we expect the observed patterns of repair to extend to other domains.
\section*{Acknowledgments}
This research has been funded by the \href{https://www.dfg.de/}{Deutsche Forschungsgemeinschaft} (DFG, German Research Foundation) -- \href{https://gepris.dfg.de/gepris/projekt/512393437}{CRC-1646, project no. 512393437}, project \href{https://gepris.dfg.de/gepris/projekt/537416633}{B02}.
We acknowledge the use of AI tools for language editing and code completion. No AI tools were used to generate ideas, analyses, or results. All code was reviewed, tested, and validated by the authors.

\bibliography{custom, references}
\newpage
\appendix
\section{Appendix}
\label{sec:appendix}
\subsection{Additional results}
\paragraph{Mean Accuracies for Both Turns}\label{app:meanAccuracies for both bositions}
\begin{table*}[ht]
    \centering
    \begin{tabular*}{\linewidth}{@{\extracolsep{\fill}}lcccc}
        \toprule
        & \multicolumn{2}{c}{Answerable} & \multicolumn{2}{c}{Unanswerable} \\
        \cmidrule(lr){2-3} \cmidrule(lr){4-5}
        Model & Second Turn & Fourth Turn & Second Turn & Fourth Turn \\
        \midrule
        GPT 
            & 0.97 & \textbf{0.97} & 0.41 
             & 0.40 \\
        Claude 
            & \textbf{0.98} & \textbf{0.97} & 0.42 
             & 0.48 \\
        Mistral 
            & \textit{0.85} & 0.40 & 0.40 
             & 0.45 \\
        Deepseek 
            & 0.95 & \textit{0.37} & \textit{0.18} 
             & \textbf{0.54} \\
        Phi 
            & 0.97 & 0.96 & \textbf{0.45} 
             & \textit{0.44} \\
        \bottomrule
    \end{tabular*}
    \caption{Accuracies of model answers across the second and fourth turn for the answerable and unanswerable subsets. Best performance per subset is marked in \textbf{bold}, weakest performance in \textit{italic}.}
    \label{tab:accuracies_both_iterations}
\end{table*}

On the right-hand side of Table \ref{tab:accuracies_both_iterations}, mean accuracies for answers in fourth turn to the three clarification request strategies are reported. On the answerable subset, Claude’s performance decreases by one percentage point compared to the second turn, whereas GPT remains unchanged and emerges as the best-performing model together with Claude, both achieving an accuracy of 97\%. Phi also shows a slight decrease of one percentage point, resulting in an accuracy of 96\%, while maintaining a high overall performance level. In contrast, the performance of Mistral and DeepSeek drops substantially, reaching accuracies of 40\% and 37\%, respectively.
For the unanswerable subset, the opposite trend is observed. Phi and GPT exhibit only a marginal decrease of one percentage point, resulting in accuracies of 44\% for Phi and 40\% for GPT. Claude, Mistral, and DeepSeek, by contrast, show notable improvements, attaining accuracies of 48\%, 45\%, and 54\%, respectively. Notably, DeepSeek, which showed comparatively low performance in the second turn, achieves the highest accuracy on the unanswerable subset in the fourth turn.
\FloatBarrier

\begin{figure}[H]
    \includegraphics[width=\columnwidth]{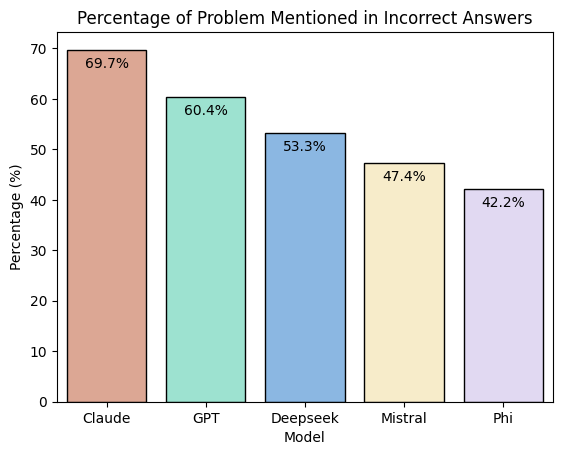}
\caption{Proportion of incorrect responses to unanswerable questions containing an explicit problem mention.}
    \label{fig:First_postition_repair}
\end{figure}

\FloatBarrier

\newpage
\subsection{Examples}
\label{sec:Excluded questions}

\begin{table*}
\caption{Overview of the reasons for excluding questions, accompanied by an example for each reason.}
\centering
\small
\begin{tabularx}{\linewidth}{@{}p{0.22\linewidth} p{0.60\linewidth} p{0.12\linewidth}@{}}
\toprule
\textbf{Reason} & \textbf{Question (with ID)} & \textbf{Gold Answer} \\
\midrule

Unit ambiguity (minutes vs.\ hours) &
\textbf{ID 5077} — Dane eats lunch at 12:30. It takes him 20 minutes to eat. What time does Dane finish eating? &
12.0 \\
\midrule

Problem yields more than one outputs, but the gold answer provides only one &
\textbf{ID 5099} — The angles of a triangle are in the ratio 1:2:3. Find each angle. &
30.0 \\
\midrule

Representation of ''more`` in the answers (multiple valid formulations) led to textual descriptions and could be mistaken for incorrect responses &
\textbf{ID 21} — 28 children were riding on the bus. At the bus stop 82 children got on the bus while some got off the bus. Then there were 30 children altogether on the bus. How many more children got on the bus than those that got off? &
2.0 \\
\bottomrule
\end{tabularx}

\label{tab:appendix_example_excluded_questions}
\end{table*}

\FloatBarrier

\label{sec:annotationexample}
\begin{table*}[t]
\caption{Examples of model answers to question \textbf{2529:} \textit{Faye was placing her pencils and crayons into 16 rows with -6 crayons and -21 pencils in each row. How many crayons does she have?} and annotations. Mentions of the trouble source are highlighted in yellow.}
\centering
\small
\begin{tabularx}{\linewidth}{@{}p{0.08\linewidth} p{0.7\linewidth}p{0.17\linewidth}@{}}
\toprule
 \textbf{Model} &\textbf{Modelanswer} & \textbf{Label} \\
 \midrule
 \textbf{Claude} & 
 I need to find the total number of crayons Faye has. 

Given information:

- 16 rows

- -6 crayons in each row

- -21 pencils in each row (this doesn't affect the crayon count)

To find the total number of crayons:

Total crayons = Number of rows × Crayons per row

Total crayons = 16 × (-6)

Total crayons = -9

\textbackslash boxed\{-96\} 
& Annotators agreed on \textit{Problem not adressed} \\
\midrule
\textbf{Deepseek} & 
Faye is arranging her pencils and crayons into 16 rows, with each row containing -6 crayons and -21 pencils. To determine the number of crayons she has, we need to consider the given numbers and their interpretation.

1. \colorbox{yellow!60}{Each row has -6 crayons, which could be interpreted as 6 crayons due to the context} \colorbox{yellow!60}{of physical objects not allowing negative counts.}

2. There are 16 rows, so we multiply the number of crayons per row by the number of rows.

Thus, the total number of crayons is calculated as follows:

- Number of crayons per row: 6

- Number of rows: 16

- Total number of crayons:  6 \textbackslash times 16 = 96 

Therefore, the final answer is \textbackslash boxed\{96\}.
& Annotators Disagreed $\rightarrow$ \textit{Problem not adressed}\\

\midrule

\textbf{GPT} & Let's understand the problem carefully. Faye is placing crayons and pencils into 16 rows. She has \(-6\) crayons and \(-21\) pencils per row. \colorbox{yellow!60}{The negative numbers suggest an error in the problem statement, as we cannot have a} \colorbox{yellow!60}{negative count of crayons or pencils. However, we'll interpret the numbers as positive} \colorbox{yellow!60}{quantities to solve the problem.}

Assuming that Faye has 6 crayons and 21 pencils per row, we'll calculate how many crayons she has in total. 

Given:

- Number of crayons per row = 6

- Number of rows = 16

The total number of crayons is the product of the number of crayons per row and the number of rows:

6 \textbackslash times 16 = 96

Thus, Faye has \textbackslash boxed\{96\} crayons. & Annotators Agreed on \textit{Problem adressed}\\

\bottomrule
\end{tabularx}

\label{tab:appendix_example_model_answers}
\end{table*}

\FloatBarrier

\subsection{Example for multiturn Prompting via Openrouter} \label{sec:codeexample}

\begin{lstlisting}[language=Python, caption={Example Python Code}]
from openai import OpenAI
client = OpenAI(
  base_url="https://openrouter.ai/api/v1",
  api_key=<API-Key>,
)

completion = client.chat.completions.create(
      model="deepseek/deepseek-r1-distill-llama-70b",
      messages=[
      {'role': 'user', 
        'content': 'Please put your final answer within \\boxed{}. Bryan took a look at his books and magazines. If he has 9 books and 46 magazines in each of his 10 bookshelves.How many magazines does he have in total?'},
      {'role': 'assistant',
        'content': 'Bryan has 46 magazines in each of his 10 bookshelves. To find the total number of magazines, we multiply the number of magazines per shelf by the number of shelves:\n\n\\[ 46 \\, \\text{magazines/shelf} \\times 10 \\, \\text{shelves} = 460 \\, \\text{magazines} \\]\n\nThus, the total number of magazines Bryan has is \\(\\boxed{460}\\).'},
        {'role': 'user', 
        'content': "Are you sure it's 460?"}
        ])

\end{lstlisting}
\FloatBarrier
\end{document}